# AUTOMATED RIVER GAUGE PLATE READING USING A HYBRID OBJECT DETECTION AND GENERATIVE AI FRAMEWORK IN THE LIMPOPO RIVER BASIN

Kayathri Vigneswaran
*Water Futures Data & Analytics*
*International Water Management Institute (IWMI)*
Colombo, Sri lanka
Kayathrivig@gmail.com

Hugo Retief
Association for Water and Rural Development (AWARD)
H.Retief@cgiar.org

Mariangel Garcia Andarcia
*Water Futures Data & Analytics*
*International Water Management Institute (IWMI)*
Colombo, Sri lanka
M.GarciaAndarcia@cgiar.org

Hansaka Tennakoon
*Water Futures Data & Analytics*
*International Water Management Institute (IWMI)*
Colombo, Sri lanka
H.Tennakoon@cgiar.org

Jai Clifford-Holmes
Association for Water and Rural Development (AWARD)
jai@award.org.za

**Accurate and continuous monitoring of river water levels is essential for flood forecasting, water resource management, and ecological protection. Traditional hydrological observation methods are often limited by manual measurement errors and environmental constraints. This study presents a hybrid framework integrating vision-based waterline detection, YOLOv8-Pose scale extraction, and large multimodal language models (GPT-4o and Gemini 2.0 Flash) for automated river gauge plate reading. The methodology involves sequential stages of image preprocessing, annotation, waterline detection, scale gap estimation, and numeric reading extraction. Experiments demonstrate that waterline detection achieved high precision (94.24%) and F1-score (83.64%), while scale gap detection provided accurate geometric calibration for subsequent reading extraction. Incorporating scale gap metadata substantially improved the predictive performance of LLMs, with Gemini Stage 2 achieving the highest accuracy (MAE = 5.43 cm, RMSE = 8.58 cm, $R^2$ = 0.84) under optimal image conditions. Results highlight the sensitivity of LLMs to image quality, with degraded images producing higher errors, and underscore the importance of combining geometric metadata with multimodal AI for robust water-level estimation. Overall, the proposed approach offers a scalable, efficient, and reliable solution for automated hydrological monitoring, demonstrating potential for real-time river gauge digitization and improved water resource management.**

## I. INTRODUCTION

Accurate acquisition of hydrological data, particularly during high flood seasons, is essential for flood forecasting and urban waterlogging alerts. Among these parameters, the water discharge is a fundamental indicator in rivers, lakes, and reservoirs, providing the basis for estimating runoff, flood discharge, sediment transport, and nutrient dynamics [1] Continuous and reliable monitoring of water discharge is therefore a critical aspect of hydrological observation, supporting disaster risk assessment, flood warnings, and water resources planning. It also plays an important role in managing public and industrial water supply, irrigation, and ecological protection. In hydropower production, rainfall, inflows, and discharge must be monitored to maximize energy generation while ensuring dam safety. Overall, sustained water-level monitoring is vital for water security, efficient resource management, and safeguarding aquatic environments [1], [2], [3].

Traditional manual gauge reading relies on visual observation, but it suffers from low measurement frequency and significant subjectivity. The accuracy is often influenced by the observer's angle and becomes highly uncertain under harsh environmental conditions. On the other hand, river discharge, although one of the most accurately measurable components of the hydrological cycle, remains difficult to access consistently. Monitoring networks are sparse across much of the globe, and no standardized mechanism exists for real-time global collection and distribution of discharge data. While some hydro-meteorological agencies (USGS, R-ArcticNet, LBA-Hydronet, etc.) have begun releasing discharge information online, variations in data formats and limited coverage hinder widespread utilization [2],[4].

Different types of water-level gauges, include ing pressure, encoder/float, ultrasonic, and radar sensors, have been developed and applied under varying hydrological conditions

[5], [6]. Water levels can be monitored using a range of sensor technologies, each with distinct performance characteristics and operational trade-offs. Common approaches include pressure transducers, encoder/float systems, ultrasonic sensors, and radar sensors. Pressure-type and encoder/float systems are relatively mature, low-power technologies that provide centimetre-level accuracy but are sensitive to water density and typically require frequent recalibration (Donegan and White, 1984; Bera et al., 2006; Guilin et al., 2007; Liu et al., 2005; Jia et al., 2009). More recent non-contact methods, such as ultrasonic and radar sensors, offer high accuracy and reduced maintenance, though their performance can be affected by atmospheric conditions (e.g., temperature, humidity, rainfall) and, in the case of radar, higher instrumentation costs (Jun and Hui, 2008; Yin et al., 2008; B. Jerry et al., 2006; Terao et al., 2007). Together, these methods illustrate the balance between cost, accuracy, robustness, and ease of deployment.

Video-based techniques for surface velocity and discharge estimation have advanced considerably in recent decades, providing non-contact alternatives to conventional in-situ measurements. Particle-based methods, including Particle Image Velocimetry (PIV) and Particle Tracking Velocimetry (PTV), rely on tracer particles to estimate instantaneous flow fields. Large-Scale PIV (LSPIV) was later adapted to enable these measurements in complex, real-world river environments [7], [8]. To address limitations associated with tracer availability and environmental noise, Spatio-Temporal Image Velocimetry (STIV) was developed, relying on the analysis of surface textures such as ripples and corrugations, to derive one-dimensional velocity profiles with high computational efficiency [9]. More recently, deep-learning-based methods, particularly convolutional neural networks, have demonstrated strong potential for automated feature extraction and flow estimation, offering improved adaptability and robustness when trained on diverse datasets [10], [11]. Overall, while particle-based and optical flow methods remain widely adopted, emerging spatio-temporal and deep learning approaches are increasingly complementing or replacing them, reflecting the trend toward automated, scalable, and more accurate video-based river discharge monitoring. Parameter uncertainties in ranging models and algorithm settings, such as window size selection in STIV and empirically derived surface velocity coefficients, contribute to substantial variability in results [12]. These limitations, including storage constraints and image quality, highlight the difficulties of ensuring consistent, reliable, and accurate flow estimation under diverse field conditions.

Recent advances in computer vision have explored image-based water level detection to overcome challenges in traditional monitoring techniques as well as video-based techniques [13]. For instance, projection methods using grayscale and edge images have demonstrated accuracies of up to 1 cm, effectively mitigating issues caused by low light, glare, shadows, and artificial illumination [13]. Similarly, automated extraction approaches by Eltner [14] have achieved sub-centimeter accuracy in simple scenes, although performance declines significantly in complex backgrounds. More recent developments employing deep learning, such as YOLOv5s-based detection, have improved robustness, achieving an average error of 1.38 cm under favorable daylight conditions [2]. Another study presents a method combining continuous image subtraction and a SegNet deep learning neural network for river water level measurement, achieving high accuracy with RMSE values between 0.013 m and 0.066 m and a correlation coefficient of 0.998, demonstrating improved performance with larger, more diverse, and higher-resolution training datasets. Despite its demonstrated effectiveness, this method exhibits several limitations that constrain its broader applicability. Without an explicitly defined reference maximum or calibration scale, accurate measurement cannot be guaranteed, limiting the method's adaptability to predefined gauge designs [15].

However, recent advancements in multimodal large language models (LLMs), as surveyed by Yin et al. 2024) highlight their advanced visual perception and reasoning capabilities, enabling them to integrate visual and textual information. These models can be programmed to autonomously identify calibration references such as the maximum scale on gauges, overcoming a significant challenge of previous methods. Leveraging LLMs' ability to dynamically detect reference points can thus enable the application of such models across multiple gauge plate designs with varying maximum reference levels without the need for manual recalibration.

This study investigates the integration of LLM-based generative AI with domain-specific object detection models for hydrological monitoring. By combining the visual reasoning capabilities of multimodal LLMs with robust object detection frameworks, this research seeks to develop a method for automated river gauge plate reading that is more adaptable under diverse field conditions. Unlike previous studies, which have not explored LLM-based generative AI in this context, this research addresses a critical gap by leveraging the ability of multimodal LLMs to autonomously identify reference maximum levels and other calibration features on various gauge plate types. This capability enables the proposed hybrid approach to be used across multiple gauge designs with different maximum reference levels, overcoming current limitations in accuracy and reliability, especially in complex visual environments where traditional image-processing or standalone deep learning methods are insufficient.

## II. MATERIALS AND SOURCES

### A. Study area

The Limpopo River Basin is a major transboundary river system in southern Africa, spanning approximately 415,000 km² across four countries: South Africa, Botswana, Zimbabwe, and Mozambique. The river travels about 1,750 km from its source in South Africa's Witwatersrand region, flowing northeast and then east to the Indian Ocean. The basin hosts a dense river network with numerous tributaries across the countries, supporting important socio-economic activities such as agriculture, fishing, and wildlife conservation. Discharge monitoring stations are distributed throughout the basin, with a mix of active and inactive stations, though exact counts vary by country and river segment [17].

## B. Algorithms

### 1) YOLOv8

YOLOv8 is one of the latest generations of the "You Only Look Once" (YOLO) object detection family, building upon the capabilities of earlier versions such as YOLOv5 and YOLOv7. It adopts a one-stage detection framework comprising four primary components: input, backbone, neck, and detection head as shown in Figure 1. The input stage integrates mosaic data augmentation, adaptive anchor computation, and grayscale padding, thereby enhancing the model's robustness under varying input conditions. For feature extraction, YOLOv8 employs a modified CSPDarknet53 backbone, which incorporates Cross-Stage Partial (CSP) connections to improve gradient flow and facilitate more efficient information transfer across network stages [18]. Within this backbone, convolutional and C2f modules act as the primary residual learning units. The C2f module, derived from the ELAN structure in YOLOv8, leverages bottleneck modules to

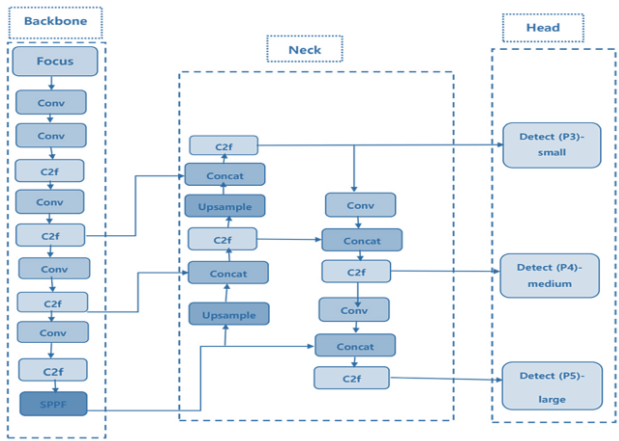

capture richer gradient information while maintaining Figure 1: YOLOV8 architecture [Source: (Ultralytics, 2025)]

computational efficiency. Additionally, the Spatial Pyramid Pooling Fast (SPPF) module expands the receptive field by combining features through pooling operations with multiple kernel sizes, further strengthening multi-scale representation [18].

The neck of YOLOv8 employs a hybrid structure combining the Feature Pyramid Network (FPN) and the Path Aggregation Network (PAN), enabling effective fusion of high-level semantic features and low-level localization details across multiple scales. This design significantly improves detection accuracy for objects of different sizes. The detection head introduces several architectural refinements, including decoupled classification and regression branches, dynamic anchor assignment through the Task Aligned Assigner, and advanced loss functions. Classification is optimized using Binary Cross-Entropy (BCE) loss, while bounding box regression is refined through Distribution Focal Loss (DFL) and Complete Intersection over Union (CIoU) loss. Prediction boxes are generated by simultaneously optimizing classification confidence and localization accuracy, with task-aligned metrics ensuring suppression of low-quality predictions. Furthermore, YOLOv8 incorporates enhanced labeling tools that support automated and customizable annotation, simplifying dataset preparation for training. Collectively, these advancements provide YOLOv8 with superior inference speed, precise bounding box estimation, and robust detection performance in complex environments characterized by overlapping objects, variable illumination, and multi-scale challenges.

### 2) YOLOV8 POSE

The YOLOv8 Pose architecture (Figure 2) was adopted to detect scale gaps on river gauge plates through keypoint localization. The model integrates object detection and pose estimation within a single framework, enabling simultaneous prediction of bounding boxes, object classes, confidence scores, and key point coordinates. Its architecture is structured into three major components: backbone, neck, and head.

The backbone serves as the feature extractor and is composed of convolutional and C2f units. Convolutional operations apply two-dimensional filtering, followed by batch normalization and a Sigmoid Linear Unit (SiLU) activation function. The C2f unit incorporates multiple convolutional blocks, with its depth determined by the model's depth multiplier (d). Depending on this multiplier, bottleneck blocks are repeated $n = 3 \times d$ or $n = 6 \times d$ times. The backbone concludes with a Spatial Pyramid Pooling Fast (SPPF) module, which aggregates multi-scale contextual information using max pooling and convolutional layers.

The neck fuses feature representations from different scales to improve detection robustness. Concatenation layers combine outputs from the backbone while preserving spatial resolution, and up sampling layers enhance the feature map resolution to align with multi-scale feature maps. The head comprises pose units designed to localize both objects and structural features. Each unit predicts bounding box coordinates, class probabilities, and key point positions, thereby enabling accurate detection of gauge plate markings. This joint detection–pose estimation framework ensures reliable identification of fine-scale structural elements under diverse field conditions.

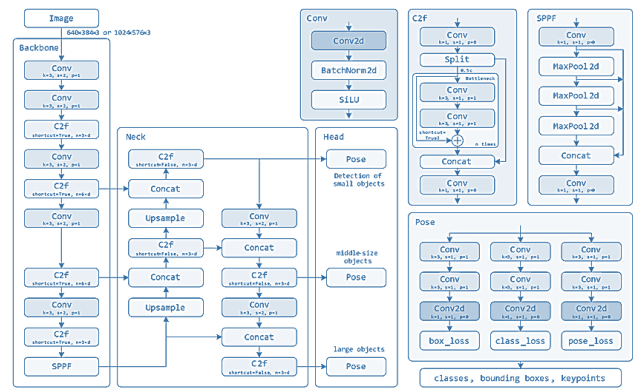

Figure 2: YOLOv8 Pose architecture [Source: (Ultralytics, 2025)]

### 3) GEMINI 2.0

Google Gemini 2.0 is Google's latest multimodal large language models engineered for joint text and image understanding. It builds on earlier Gemini versions (1.0, 1.5, etc.). It integrates vision language alignment with high-capacity reasoning, enabling tasks such as OCR (Optical Character Recognition), visual question answering, and fine-grained feature recognition. Its expanded context improves robustness in identifying pointers, gradations, and numeric labels from images [19].

*4) GPT-4o*

GPT-4o (Generative Pre-trained Transformer 4 Omni) is a multimodal large language model developed by OpenAI that processes both visual and textual inputs in a unified framework. Unlike traditional OCR models that rely solely on pixel-level text extraction, GPT-4o integrates visual perception with contextual reasoning, enabling it to interpret numeric information [20].

Data Sources

The dataset used in this study comprises 548 field-captured images of river gauge plates, many of which include surrounding environmental context rather than focusing solely on the scale. The image file sizes range from approximately 55 KB to 520 KB, with dimensions varying between 597 × 1280 pixels and 792 × 1408 pixels. The quality of the images is highly diverse, encompassing both clear, well-illuminated, unobstructed images with visible gauge markings, as well as more challenging cases affected by noise, blur, shadows, corrosion, poor contrast, and partial occlusions. This variability reflects the real-world conditions under which gauge plates are photographed, offering a representative dataset for evaluating the robustness and accuracy of automated visual water level extraction methods.

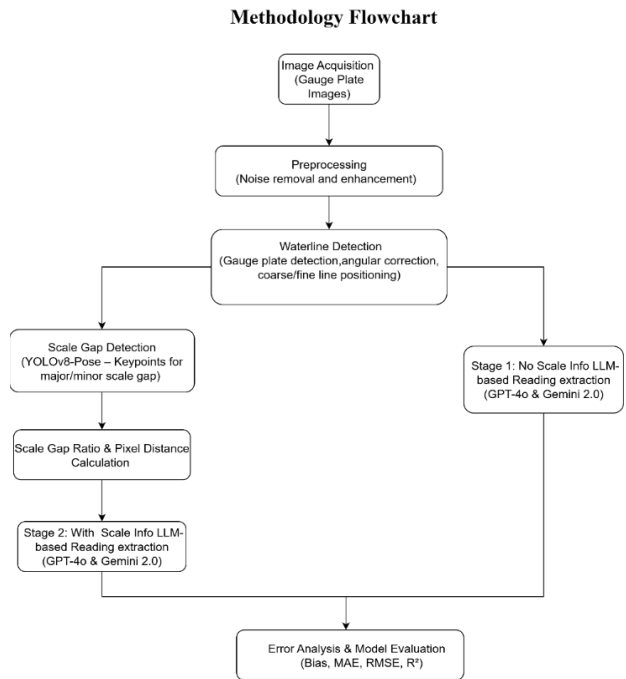

Figure 3: Methodology Flowchart [Source: IWMI]

III. METHODOLOGY

The proposed framework integrates vision-based image processing, object detection, and large multimodal models (LLMs) to automate water-level estimation from river gauge plate imagery. The methodology comprises six key stages: preprocessing, annotation, waterline detection, scale gap ratio estimation, and reading extraction using GPT and Gemini model and error matrix calculation (Figure 3).

*A. Data preparation / Dataset splitting*

The available gauge plate images were divided into training and testing sets to evaluate the model's performance. 80% of the images were used for training the AI model, while the remaining 20% were reserved for testing. This split ensures the model is trained on a majority of the data while retaining an independent set to assess generalization.

*B. Preprocessing*

*1) Noise reduction and Image enhancement*

The image dataset used for river gauge plate analysis often contains significant noise, which can adversely affect accurate water-level detection. To address this, image preprocessing techniques were employed to effectively reduce noise, enhance contrast, while preserving critical edge information. Initially, all images were loaded in color and converted to grayscale, standardizing illumination differences between daytime and nighttime images and reducing computational complexity. Grayscale conversion (standard luminosity method) for also emphasizes intensity variations, which are critical for detecting gauge edges. Gaussian smoothing was applied to the grayscale images to reduce general noise while preserving overall structural information, ensures that high-frequency noise is attenuated without significantly blurring the edges of key features.

Apart of noise reduction was achieved using a median filter with a 3×3 kernel applied pixel by pixel. This filter smooths the image while retaining important edge details, in contrast to average filtering, which can blur boundaries [13]. The median filter is mathematically defined as Eq. 1 :

$$R = \frac{1}{MxN} \sum_{i=0}^{M-1} \sum_{j=0}^{N-1} z_{i,j} \qquad (Eq:1)$$

where $M$ and $N$ define the dimensions of the filter window and $z_{i,j}$ represents the pixel intensity at position (i,j). Following denoising, images were binarized using Otsu's thresholding method, which automatically determines an optimal threshold to separate foreground (gauge plate) from the background.

*2) Annotation*

Images were annotated using Roboflow, where objects of interest were marked with rectangular bounding boxes. The coordinates of each box were normalized relative to image dimensions, as required by YOLOv8 and YOLOv8 Pose. Annotations were exported in required formats, containing class labels and bounding box information. These labeled images served as ground truth for training the object detection model.

*C. Water line detection*

*1) Angular correction*

Images of river gauges often suffer from angular misalignment due to variations in camera placement or perspective distortion. Such misalignment can introduce significant errors in automated detection of the water line, making angular correction a necessary preprocessing step before water line extraction. Gradient information was then extracted using the Sobel operator in the horizontal direction (Eq. 2), which highlights strong vertical features corresponding to gauge markings:

$$G_x = \frac{\partial I}{\partial x} \approx \begin{bmatrix} -1 & 0 & +1 \\ -2 & 0 & +2 \\ -1 & 0 & +1 \end{bmatrix} * I \qquad (Eq: 2)$$

where $I$ denotes the grayscale image and $*$ represents convolution.

A binary edge map was generated through thresholding, followed by probabilistic Hough Transform to detect line segments. For each detected line, the orientation angle θ was computed with respect to the horizontal axis (Eq. 3):

$$\theta = arctan\left(\frac{y_2 - y_1}{x_2 - x_1}\right) \qquad (Eq: 3)$$

where $(x_1, y_1)$ and $(x_2, y_2)$ are the endpoints of the line segment. To ensure robustness against outliers and spurious detections, only sufficiently long line segments (>150 pixels) were considered. The distribution of line orientations was then analyzed, and the 80th percentile angle together with the mean absolute angle were used to determine the dominant skew (Figure 4).

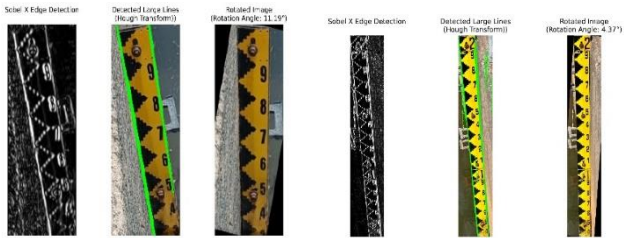

Figure 5 : Angular correction followed by edge detection and Hough transform [Source: IWMI]

*2) Coarse line positioning*
water line detection was performed in two steps: coarse and fine positioning as shown in Figure 5. In the coarse step, the grayscale image was enhanced, and edges were extracted using the Canny operator [13]. The horizontal edge intensity profile was computed by summing edge magnitudes along each row, and the row with maximum change was selected as the coarse water line.

*3) Fine line positioning*
For fine positioning, vertical gradient (Sobel Y) was computed within a narrow region (± 5 rows) centered on coarse line [13]. The row exhibiting the maximum vertical gradient was identified as the precise water line. This two-stage approach, combining angular correction with coarse-to-fine water line localization, ensures accurate and consistent detection.

*D. Scale gap ratio estimation*
The water level on a river gauge plate was determined by integrating information from both major and minor scales.

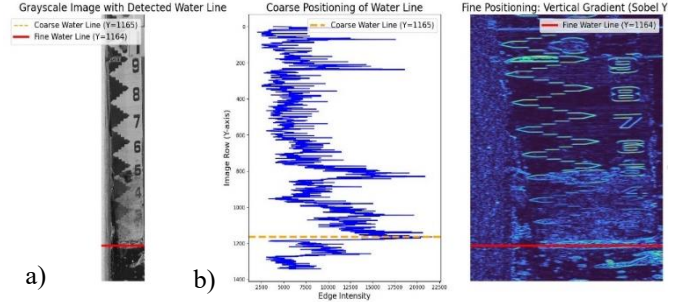

Figure 4 : Waterline detection (a) gray scale transformation (b) coarse position (c) fine position [Source: IWMI]

The **major scale** is represented by bold numerals at fixed intervals of 10 cm, while the **minor scale** consists of triangular divisions located between successive major scales, enabling finer precision in water-level estimation as in Figure 6. YOLOv8 pose was used to detect the major scale gap, which is featured by key point detection technique. To standardize measurements, the **major scale gap** ($D_m$) was computed as the median of the differences between consecutive major scale points Eq.4:

$D_m$ = Median ($s_2 - s_1, s_3 - s_2, s_4 - s_3,....., s_n - s_{(n-1)}$ )    (Eq: 4)

At the waterline intersection, scale gap at the waterline ($D_n$) was defined as Eq.5:

$D_n = s - s_0$    (Eq: 5)

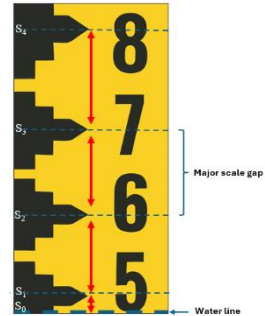

Figure 6: Diagram of template gauge plate used in the field [Source: IWMI]

To achieve sub-interval precision, the ratio of the **major scale gap at the waterline** to the major scale gap (R) was calculated. This ratio is given by:

$$R = \frac{D_n}{D_m} = \frac{\text{Real reading corresponding to } D_n}{10} \qquad (Eq: 6)$$

Finally, the **water line reading** (W) was obtained by combining the contributions of both $D_n$ and $D_m$:

$$W = M - (R*10) \qquad (Eq: 7)$$

Where M is the major scale reading at the water line, R is the calculated scale ratio. This formulation allows the framework to achieve precise water-level estimation by leveraging geometric relationships between major and minor scale divisions.

*E. Model assessment*

The performance of the trained YOLOv8 pose model in detecting scale gap was evaluated using standard detection metrics, including precision, recall, Mean Average Precision (MAP) at Intersection over Union (IoU) threshold 0.5 (MAP0.5) from the PASCAL VOC metric, and MAP0.5–0.95 from the COCO metric. The evaluation was conducted at an input resolution of $640 \times 640$ px.

*F. Reading extraction using LLMs*

Waterline readings from river gauge plates were extracted using multimodal LLMs, GPT-4o and Gemini 2.0 Flash, capable of interpreting numeric values directly from images. The procedure was implemented in two stages:

**Stage 1:** The models received only the preprocessed, waterline-detected images, requiring them to infer both the gauge reading and the scale spacing.
**Stage 2:** The models were provided with additional metadata, specifically the scale gap ratio, allowing accurate conversion of pixel distances to real-world measurements.

Images were input into the models with structured prompts, and then the numeric waterline readings were extracted. A structured prompt was designed to guide the AI model in extracting water-level readings from cropped river gauge plate images. Each image was pre-processed so the lowest pixel corresponds to the water–level intersection. Since the gauge plate follows a 1-cm resolution with digits in a sequential order (1,2,3…), the prompt instructs the model to identify the topmost visible digit, determine the full number sequence from top to bottom, and detect any partially visible digit at the lowest edge. The scale gap ratio algorithm was integrated with the above extraction to get the final waterline reading

*G. Outlier removal and **Error Metrics***

To ensure robust and reliable water-level estimation, extreme or anomalous predictions were identified and removed using an Interquartile Range (IQR) based method[21]. The performance of water-level predictions was quantitatively evaluated using multiple complementary metrics: **bias, Mean Absolute Error (MAE), Root Mean Squared Error (RMSE), $R^2$,** as well as **precision, recall, and F1-score** derived from the confusion matrix.

**Bias:** measures the average deviation of predicted readings from ground-truth values, indicating systematic over or underestimation.
**MAE:** calculates the mean absolute difference between predicted and observed readings, reflecting overall prediction accuracy.

**RMSE:** emphasizes larger deviations by considering the square root of the average squared errors, capturing variability in predictions.
**$R^2$** represents the proportion of variance in observed readings explained by the predictions, indicating goodness of fit.
**Confusion matrix-based metrics**: A prediction was considered a True Positive if the detected waterline fell within a ($\pm 5$) pixel row zone around the annotated line. **Precision** quantifies the fraction of correct positive predictions, **recall** represents the fraction of actual positives correctly detected, and the **F1-score** provides a harmonic mean of precision and recall, offering a balanced measure of detection performance.

This comprehensive suite of metrics allows for robust assessment of model accuracy, reliability, and consistency across different experimental stages and image quality categories.

## IV. RESULTS

*A. Experimental Environment*

The experiments in this study were conducted on a system running Windows 10 with a 13th Gen Intel(R) Core(TM) i7-1355U processor @ 1.70 GHz, 16 GB RAM (15.6 GB usable), and a 64-bit x64-based architecture. The system includes an integrated GPU with 128 MB memory; however, model training performed more efficiently on the CPU. Object detection and keypoint estimation tasks were implemented using YOLOv8 and YOLOv8-Pose, developed on the PyTorch deep learning framework. The average training times were 14.67 minutes for YOLOv8 and 87.27 minutes for YOLOv8-Pose. Reading extraction was done using Google Gemini 2.0 as well as GPT 4o MINI as a comparative analysis.

*B. Waterline detection performance*

The developed framework effectively identified waterline positions across diverse river gauge images. Figure 7 presents examples of the detected waterlines superimposed on the original images, demonstrating the robustness of the pipeline under varying visual conditions. The task involved sequential stages of object detection, angular correction, coarse line estimation, and fine line refinement.

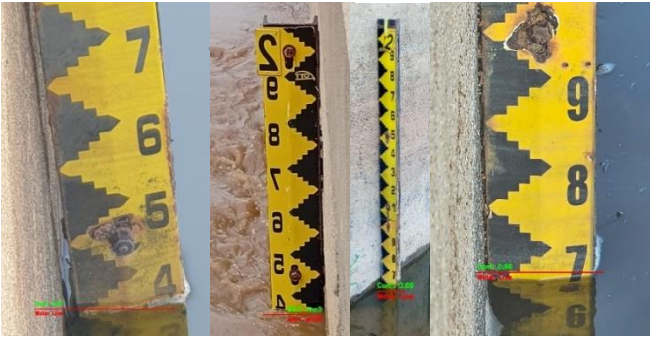

Figure 8 : images of gauge plate where water line (red line) is detected, the confidence level of detection is also mentioned (green text). [Source: IWMI]

Model evaluation was performed using a zone of (± 5) pixel rows around the annotated waterline. A detection was considered a True Positive when it intersected this zone. To reduce false detections, a confidence threshold of 0.20 was applied, resulting in the rejection of 5.76% of images across all quality categories. This threshold was used to improve the quality of results in terms of rejecting false positives. The confusion matrix which shows the results of images confidence > 0.2 (Figure 8) remained nearly perfect, confirming that most predictions were accurately aligned with the annotated reference.

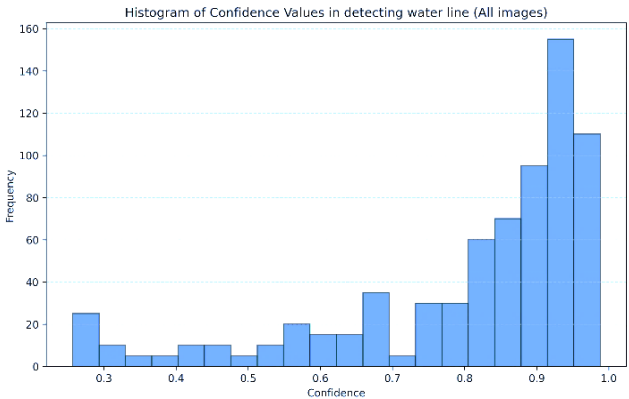

Figure 9: confidence distribution on water line detection where threshold >0.2 [Source: IWMI]

Quantitative assessment showed an overall precision of 94.24%, with F1-score of 83.64%, derived at confidence threshold of 0.20. These results highlight the model's capacity to achieve high precision while maintaining a balanced recall. The confidence distribution (Figure 9) further substantiates the reliability of the detections, with the majority of predictions concentrated in the high-confidence range (0.85–1.0). Lower confidence values (0.30–0.60) were observed in more challenging cases, primarily caused by scale invisibility due to corrosion, interference from surrounding objects such as grass and debris, and image blurring. Overall, the model demonstrated strong performance, consistently detecting waterlines with high reliability and accuracy, even in degraded image conditions.

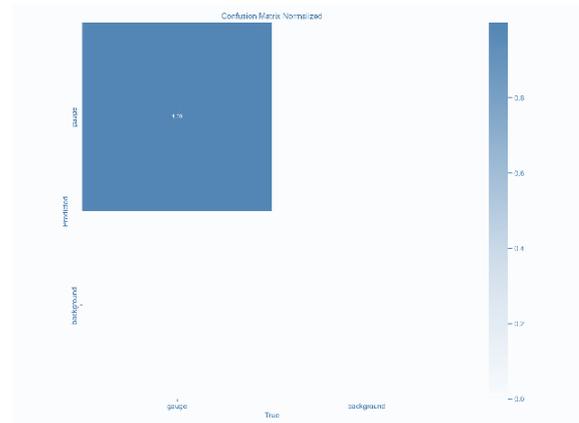

Figure 7: Distribution of all prediction where the threshold >0.2 which poses zero false positive [Source: IWMI]

C. Scale gap detection performance

The results demonstrate that the model achieves high detection accuracy for major scale gaps on the gauge plate, with a precision of 0.81%, recall of 0.87%, MAP0.5 of 0.89%, and MAP0.5–0.95 of 0.80% (Table 1).. Training at this resolution required approximately 162.56 minutes.

Table 1: Error metric summary on scale gap detection [Source: IWMI]

| METRIC | MAJOR SCALE |
|---|---|
| **PRECISION (%)** | 0.81 |
| **RECALL (%)** | 0.87 |
| **MAP50 (%)** | 0.89 |
| **MAP50-95 (%)** | 0.80 |

In addition to these performance indicators, the model's outputs generated essential parameters for subsequent computation of water-level readings such as Major scale gap in pixels (Dm), distance from the bottom of the gauge plate to the waterline (Dn) and Total pixel height of the gauge plate (Hy) which is input into LLM for reading extraction.

D. Effects of GPT/Gemini integration

Recent multimodal large language models, such as GPT-4o and Gemini 2.0 Flash, have demonstrated significant potential for optical character recognition (OCR) in hydrological monitoring. By combining visual perception with language understanding, these models move beyond traditional OCR engines, enabling reliable extraction of numeric readings from river gauge plates even under noisy, distorted, or low-quality conditions. GPT-4o provides strong context-aware accuracy for complex scenarios, while Gemini 2.0 Flash emphasizes high-speed, scalable processing, making them complementary solutions for accurate and efficient water-level digitization. Both GPT-4o and Gemini 2.0 Flash were applied for water-level reading extraction from gauge plate imagery, allowing for a comparative evaluation of their accuracy, adaptability, and performance under diverse field conditions [22].

In Stage 1, GPT achieved an $R^2$ of 0.13 and Gemini 0.16, indicating weak correlations and highlighting the difficulty of inferring both waterline position and scale spacing solely from the images.

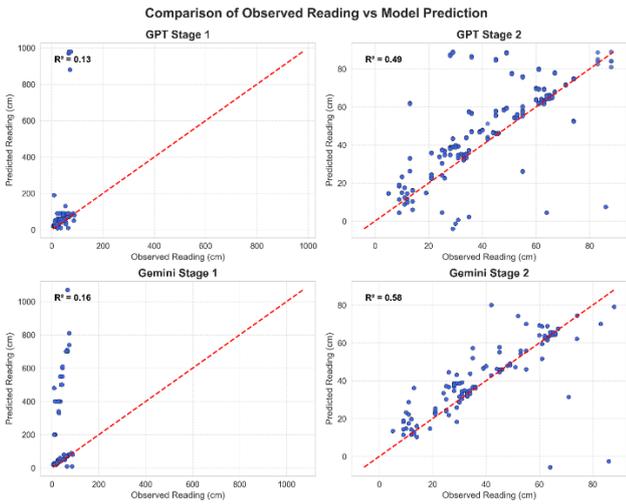

Figure 10: trend comparison of observed reading vs model prediction (before removal of outliers) [Source: IWMI]

In Stage 2, the models improved in performance substantially, with GPT reaching an R² of 0.49 and Gemini 0.58. Scatter plots (Figure 10) show better alignment with ground truth in Stage 2, demonstrating that integrating scale gap information significantly enhances predictive accuracy. These results underscore the value of combining multimodal LLMs with domain-specific metadata for robust and precise water-level estimation from gauge plate imagery.

After outlier removal, the scatter plots were reconstructed to assess the improvement in prediction accuracy as in Figure 11. For GPT Stage 1, 5.04% of outliers were removed, increasing the R² from 0.13 to 0.45, while GPT Stage 2 achieved an R² of 0.49 with no outliers, showing a moderate improvement when scale gap information was included. In contrast, Gemini Stage 1 initially exhibited a high R² of 0.63, but this required removing 24.37% of extreme predictions, indicating a substantial number of outlier values. With Stage 2, Gemini achieved an R² of 0.58 with zero outliers, demonstrating robust performance without needing any extreme value removal. These results highlight that including scale gap ratio metadata enhances model accuracy, and although Gemini can achieve higher R², it is more sensitive to outliers in the absence of scale information, whereas GPT shows consistent improvement with scale gap integration.

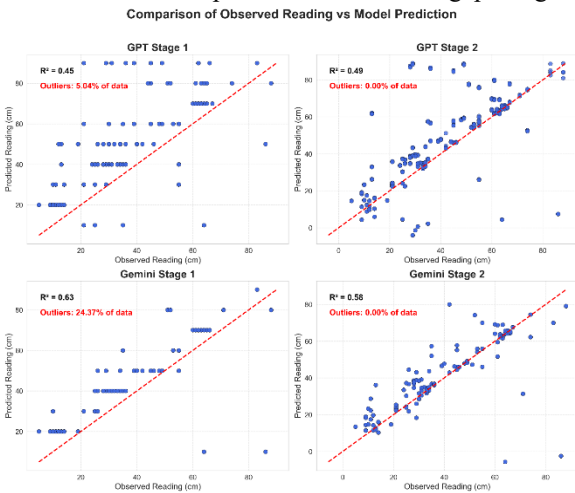

Figure 11 : trend comparison of observed reading vs model prediction (after removal of outliers) [Source: IWMI]

The error metrics of water-level predictions are summarized in Table 2. In GPT Stage 1, the model exhibited a bias of 13.97 cm, MAE of 17.35 cm, and RMSE of 22.58 cm, reflecting substantial overestimation and high variability in predictions. The inclusion of the scale gap ratio in GPT Stage 2 markedly improved performance, reducing the bias to 4.34 cm, MAE to 9.99 cm, and RMSE to 17.49 cm, demonstrating enhanced accuracy and precision.

Table 2: Model performance metrics after removal of outliers [Source: IWMI]

| MODEL | BIAS (CM) | MAE (CM) | RMSE (CM) |
|---|---|---|---|
| GPT STAGE 1 | 13.97 | 17.35 | 22.58 |
| GPT STAGE 2 | 4.34 | 9.99 | 17.49 |
| GEMINI STAGE 1 | 7.06 | 10.27 | 14.28 |
| GEMINI STAGE 2 | 1.98 | 6.97 | 13.68 |

For the Gemini models, Stage 1 predictions showed a bias of 7.06 cm, MAE of 10.27 cm, and RMSE of 14.28 cm. While initial performance was better than GPT Stage 1, the model exhibited sensitivity to extreme values, as indicated by the outlier analysis. Incorporating scale gap information in Gemini Stage 2 resulted in the lowest overall errors, with a bias of 1.98 cm, MAE of 6.97 cm, and RMSE of 13.68 cm, indicating highly accurate and reliable water-level estimation. These findings confirm that the integration of scale gap metadata substantially enhances the predictive performance of both models, with Gemini Stage 2 providing the most precise and robust results.

### E. Performance on Optimal Quality Images

To evaluate model performance under ideal conditions, only images with clear scales, daylight, no blur, no corrosion, and no visual obstacles were considered. The results are summarized in Table 3 and visualized using a bar chart for direct comparison (Figure 12).

Table 3: Performance matrix of optimal- quality images [Source: IWMI]

| MODEL | BIAS (CM) | MAE (CM) | RMSE (CM) | R² |
|---|---|---|---|---|
| GPT STAGE 1 | 14.71 | 16.56 | 21.68 | 0.54 |
| GPT STAGE 2 | 4.89 | 9.33 | 15.99 | 0.56 |
| GEMINI STAGE 1 | 7.90 | 9.38 | 11.71 | 0.80 |
| GEMINI STAGE 2 | 3.04 | 5.43 | 8.58 | 0.84 |

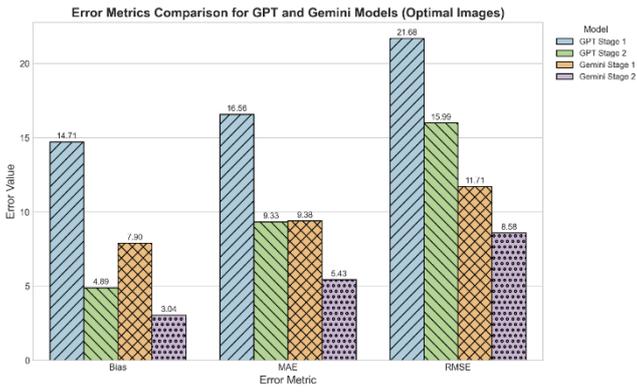

*Figure 12: trend comparison of observed reading vs model prediction (optimal images) [Source: IWMI]*

The results demonstrate that the addition of scale gap ratio information (Stage 2) consistently improves water-level estimation for both GPT and Gemini models. Gemini Stage 2 achieved the best performance, with the lowest bias and error (MAE = 5.43 cm, RMSE = 8.58 cm) and the highest R² value (0.84), indicating strong correlation with ground-truth measurements. GPT models also show improvements in Stage 2, though the errors remain higher than Gemini, highlighting the advantage of Gemini in optimal visual conditions (Figure 12).

The results from Figure 13 shows that, for the GPT model, the R² value improved from 0.45 to 0.54 in Stage 1, representing a 20% increase, and from 0.48 to 0.56 in Stage 2, indicating a 16.7% improvement. The Gemini model exhibited even stronger gains, with Stage 1 increasing from 0.60 to 0.80 (33.3% improvement) and Stage 2 from 0.58 to 0.84 (44.8% improvement). These results emphasize that good quality images significantly enhance the reliability of multimodal LLMs and Object detection models for numerical reading extraction. The higher improvement observed for the Gemini model suggests its superior ability to utilize fine-grained visual and contextual information when processing clear, well-defined gauge plate imagery.

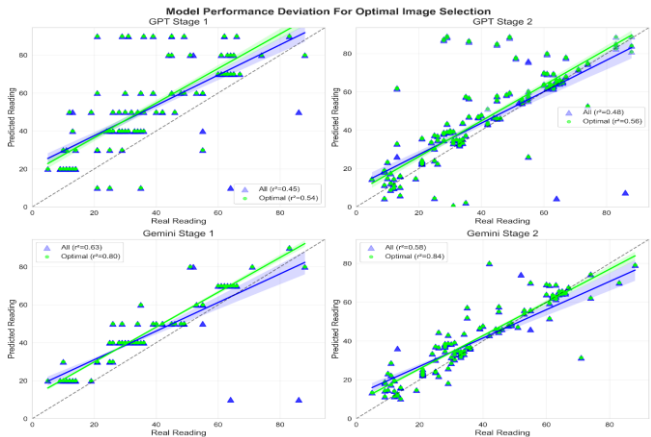

*Figure 13: Model Performance Deviation for Optimal Image Selection [Source: IWMI]*

### F. Performance on multiple image categories

For all images, Gemini Stage 2 achieved the lowest bias (1.98 cm), MAE (6.97 cm), and RMSE (13.68 cm), outperforming GPT models. When considering only optimal quality images performance improved for all models, with Gemini Stage 2 again achieving the best accuracy (bias = 3.04 cm, MAE = 5.43 cm, RMSE = 8.58 cm), highlighting the advantage of high-quality input. Conversely, sub-optimal images showed substantially higher errors, particularly for Gemini Stage 1 (bias = 193.25 cm, MAE = 195.75 cm, RMSE = 269.91 cm), indicating sensitivity to poor visual conditions. Heat maps of the error matrices (Figure 14) illustrate these trends, with optimal images exhibiting concentrated low-error regions, while sub-optimal images display widespread high-error zones, emphasizing the importance of image quality. The

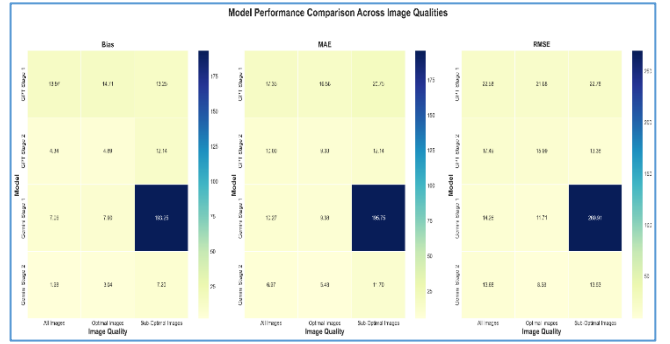

*Figure 14 : Model performance visualization heat map across all quality images [Source: IWMI]*

Table 4 represents average error generated per reading which further supports these findings, with GPT Stage 1, GPT Stage 2, Gemini Stage 1, and Gemini Stage 2 yielding mean deviations of approximately ±16.57 cm, ±9.28 cm, ±8.84 cm, and ±5.37 cm, respectively. The steady reduction in average error across model stages demonstrates the effectiveness of incorporating scale gap ratio information and the use of higher quality images.

Table 4: Average error generated per readings for optimal images [Source: IWMI]

| MODEL | AVERAGE ERROR (CM) |
| --- | --- |
| GPT STAGE 1 | ±16.57 |
| GPT STAGE 2 | ±9.28 |
| GEMINI STAGE 1 | ±8.84 |
| GEMINI STAGE 2 | ±5.37 |

## V. DISCUSSION

The proposed framework combining YOLOv8-based waterline detection with multimodal LLMs, GPT-4o and Gemini 2.0 Flash, demonstrated robust and accurate river gauge plate reading. Waterline detection achieved high precision (94.24%) and F1-score (83.64%), even under challenging visual conditions, validating the reliability of the pipeline for subsequent scale-based water-level estimation. The use of a positive zone and confidence threshold effectively minimized false positives, while confidence distributions highlighted consistent detection performance across diverse image qualities.

Scale gap detection using YOLOv8-Pose accurately extracted major scale gaps, waterline intersections, and total plate height, enabling precise minor-to-major scale ratio computations. Integrating this geometric information with GPT and Gemini substantially improved reading extraction. Stage 1 models, which lacked scale metadata, exhibited weak correlations with ground truth ($R^2$: GPT = 0.13, Gemini = 0.16), whereas Stage 2 models, incorporating scale gap ratios, showed marked improvements ($R^2$: GPT = 0.48, Gemini = 0.58). Outlier analysis indicated that Gemini Stage 1 was sensitive to extreme predictions (24.37% outliers), whereas Stage 2 mitigated this, and GPT models showed consistent improvement with scale information.

Error metrics across all image categories highlighted the influence of image quality and metadata integration. Gemini Stage 2 consistently achieved the lowest bias, MAE, and RMSE, particularly for optimal quality images (MAE = 5.43 cm, RMSE = 8.58 cm, Bias = 3.04 cm), illustrating the importance of clear scales and unobstructed visual features. Sub-optimal images led to substantially higher errors, especially for Gemini Stage 1 (MAE = 195.75 cm, RMSE = 269.91 cm, Bias = 193.25 cm), confirming the sensitivity of LLM predictions to poor visual conditions as well for the lack of scale gap-ratio information integration. Heat maps further emphasized concentrated low-error regions for high-quality images and widespread errors for degraded inputs.

The results demonstrate that Gemini Stage 2 exhibits strong and reliable interpretation capabilities for river gauge reading, with performance highly dependent on image quality. Under optimal imaging conditions characterized by clear scales, daylight, no blur, corrosion, or visual obstacles, the model achieved high accuracy with MAE = 5.43 cm, RMSE = 8.58 cm, and Bias = 3.04 cm. However, in sub-optimal conditions involving noise, blur, shadows, poor contrast, or partial occlusions, performance significantly declined to MAE = 11.70 cm, RMSE = 13.53 cm, and Bias = 7.20 cm. When evaluating the entire dataset, the aggregated errors were MAE = 6.97 cm, RMSE = 13.68 cm, and Bias = 1.96 cm. These findings indicate that while Gemini Stage 2 serves as an effective solution applicable to multiple gauge designs with varying maximum reference levels, its accuracy remains comparable to that reported by Liu and Huang (2024), who achieved RMSEs ranging from 0.013 m to 0.066 m for predefined gauge designs.

These results demonstrate that the combination of vision-based waterline detection, YOLOv8-Pose for precise scale extraction, and LLM-based reading extraction provides a robust, automated approach for river gauge monitoring. Incorporating scale gap metadata is critical for predictive accuracy, while optimal image quality further enhances performance, highlighting the potential of multimodal models for scalable and reliable hydrological applications.

## VI. CONCLUSION

This study presents a robust framework for automated river gauge monitoring by integrating vision-based waterline detection, YOLOv8-Pose for precise scale extraction, and LLM-based reading extraction using GPT-4o and Gemini 2.0 Flash. The combination of geometric scale information with multimodal AI significantly improves water-level estimation, achieving higher accuracy, lower bias, and reduced errors, particularly under optimal image conditions. Gemini Stage 2 consistently outperformed other configurations, demonstrating the benefits of combining high-quality visual input with scale metadata. Overall, the proposed hybrid approach offers a scalable, efficient, and reliable solution for hydrological monitoring, with potential applications in real-time river gauge digitization and water resource management.

## VII. FUTURE WORK

Future research will focus on improving both the technical robustness and operational scalability of the proposed framework. Enhancements in the training process, including the incorporation of a more diverse and extensive dataset covering different gauge plate types, lighting conditions, and water turbidity levels, are expected to improve model generalization and accuracy. Additionally, refining prompt engineering strategies for the LLMs (GPT and Gemini) can further enhance the precision and contextual understanding of numerical reading extraction, particularly in complex or partially obscured images.

In parallel, the integration of the optimized model into a pilot-scale application combining the vision-based waterline detection, YOLOv8-Pose scale estimation, and LLM-based reading extraction will be pursued. This deployment will serve as a proof-of-concept for real-time, automated river gauge monitoring, enabling validation under field conditions and facilitating future integration into larger hydrological monitoring systems or citizen science platforms.

## ACKNOWLEDGMENTS

This research was conducted as part of the CGIAR Initiative on Digital Innovation and the International Water Management Institute's *Digital Innovations for Water Secure Africa (DIWASA)* project, supporting the development of digital twin approaches for hydrological monitoring and decision-making. The authors gratefully acknowledge the financial support provided by CGIAR and by The Leona M. and Harry B. Helmsley Charitable Trust. Their contributions were essential to enabling the implementation and completion of this study.

## REFERENCE

[1] H. Moradkhani and S. Sorooshian, "General Review of Rainfall-Runoff Modeling: Model Calibration, Data Assimilation, and Uncertainty Analysis," in *Hydrological Modelling and the Water Cycle: Coupling the Atmospheric and Hydrological Models*, S. Sorooshian, K.-L. Hsu, E. Coppola, B. Tomassetti, M. Verdecchia, and G. Visconti, Eds., Berlin, Heidelberg: Springer, 2008, pp. 1–24. doi: 10.1007/978-3-540-77843-1_1.

[2] G. Qiao, M. Yang, and H. Wang, "A Water Level Measurement Approach Based on YOLOv5s," *Sensors*, vol. 22, no. 10, p. 3714, Jan. 2022, doi: 10.3390/s22103714.

[3] C. Xu, "Climate Change and Hydrologic Models: A Review of Existing Gaps and Recent Research Developments," *Water Resour. Manag.*, vol. 13, no. 5, pp. 369–382, Oct. 1999, doi: 10.1023/A:1008190900459.

[4] B. Fekete and C. Vörösmarty, "The current status of global river discharge monitoring and potential new technologies complementing traditional discharge measurements," *Proc. PUB Kick- Meet.*, vol. 309, pp. 20–22, Dec. 2002.

[5] M. Donegan and B. White, "PHAG - A Microprocessor Based Tide Gauge," in *OCEANS 1984*, Sept. 1984, pp. 254–258. doi: 10.1109/OCEANS.1984.1152356.

[6] Jun, M. and Hui, W., "Design of intelligent transmitter with digital retrieval water-level sensor," *Chinese Journal of Scientific Instrument*, vol. 29, no. 4, p. 740, 2008.

[7] J. Le Coz, A. Hauet, G. Pierrefeu, G. Dramais, and B. Camenen, "Performance of image-based velocimetry (LSPIV) applied to flash-flood discharge measurements in Mediterranean rivers," *J. Hydrol.*, vol. 394, no. 1, pp. 42–52, Nov. 2010, doi: 10.1016/j.jhydrol.2010.05.049.

[8] K. Ohmi and H.-Y. Li, "Particle-tracking velocimetry with new algorithms," *Meas. Sci. Technol.*, vol. 11, no. 6, p. 603, June 2000, doi: 10.1088/0957-0233/11/6/303.

[9] R. Tsubaki, "On the Texture Angle Detection Used in Space-Time Image Velocimetry (STIV)," *Water Resour. Res.*, vol. 53, no. 12, pp. 10908–10914, 2017, doi: 10.1002/2017WR021913.

[10] S. Ansari, C. D. Rennie, E. C. Jamieson, O. Seidou, and S. P. Clark, "RivQNet: Deep Learning Based River Discharge Estimation Using Close-Range Water Surface Imagery," *Water Resour. Res.*, vol. 59, no. 2, p. e2021WR031841, 2023, doi: 10.1029/2021WR031841.

[11] Wang, J., Zhang, Y, Zhang, G., Ouyang, X., and Liu, X., "A method of applying deep learning based optical flow algorithm to river flow discharge measurement," *Measurement Science and Technology*, vol. 35, no. 6, p. 065303, 2024.

[12] Liu, W.C., Huang, W.C., and Young, C.C., "Uncertainty analysis for image-based streamflow measurement: The influence of ground control points," *water*, vol. 15, no. 1, p. 123, 2022.

[13] Z. Zhang, Y. Zhou, H. Liu, L. Zhang, and H. Wang, "Visual Measurement of Water Level under Complex Illumination Conditions," *Sensors*, vol. 19, no. 19, p. 4141, Sept. 2019, doi: 10.3390/s19194141.

[14] A. Eltner, M. Elias, H. Sardemann, and D. Spieler, "Automatic Image-Based Water Stage Measurement for Long-Term Observations in Ungauged Catchments," *Water Resour. Res.*, vol. 54, no. 12, p. 10,362-10,371, 2018, doi: 10.1029/2018WR023913.

[15] W.-C. Liu and W.-C. Huang, "Evaluation of deep learning computer vision for water level measurements in rivers," *Heliyon*, vol. 10, no. 4, p. e25989, Feb. 2024, doi: 10.1016/j.heliyon.2024.e25989.

[16] S. Yin *et al.*, "A survey on multimodal large language models," *Natl. Sci. Rev.*, vol. 11, no. 12, p. nwae403, Nov. 2024, doi: 10.1093/nsr/nwae403.

[17] P. Trambauer, M. Werner, H. C. Winsemius, S. Maskey, E. Dutra, and S. Uhlenbrook, "Hydrological drought forecasting and skill assessment for the Limpopo River basin, southern Africa," *Hydrol. Earth Syst. Sci.*, vol. 19, no. 4, pp. 1695–1711, Apr. 2015, doi: 10.5194/hess-19-1695-2015.

[18] Ultralytics, *Ultralytics YOLO*. (May 11, 2025). Accessed: May 11, 2025. [Online]. Available: https://github.com/ultralytics/ultralytics/tree/main

[19] A. G, G. D. Rao, P. S. R. Gopal, K. M, and B. S. V. Vignesh, "Automated Document Processing: Combining OCR and Generative AI for Efficient Text Extraction and Summarization," in *2024 International Conference on Innovative Computing, Intelligent Communication and Smart Electrical Systems (ICSES)*, Chennai, India: IEEE, Dec. 2024, pp. 1–5. doi: 10.1109/ICSES63760.2024.10910510.

[20] Y. Shi *et al.*, "Exploring OCR Capabilities of GPT-4V(ision) : A Quantitative and In-depth Evaluation," Oct. 29, 2023, *arXiv*: arXiv:2310.16809. doi: 10.48550/arXiv.2310.16809.

[21] C. Zhao and J. Yang, "A Robust Skewed Boxplot for Detecting Outliers in Rainfall Observations in Real-Time Flood Forecasting," *Adv. Meteorol.*, vol. 2019, no. 1, p. 1795673, 2019, doi: 10.1155/2019/1795673.

[22] Sergey Filimonov, "Ingesting Millions of PDFs and why Gemini 2.0 Changes Everything," Ingesting Millions of PDFs and why Gemini 2.0 Changes Everything. Accessed: July 07, 2025. [Online]. Available: https://www.sergey.fyi/articles/gemini-flash-2